\newif\ifdraft
\draftfalse

\PassOptionsToPackage{table}{xcolor}

\documentclass[lettersize,journal]{IEEEtran}

\usepackage{xcolor}  
\usepackage{cite}
\usepackage{amsmath,amssymb,amsfonts}
\usepackage{bbm}
\usepackage{graphicx}

\usepackage{hyperref}
\usepackage{caption,subcaption}
\usepackage{multirow}
\usepackage{cleveref}
\usepackage{comment}
\usepackage{hhline}
\usepackage{caption,subcaption}

\usepackage{siunitx}
\usepackage{booktabs}
\usepackage{csvsimple}

\usepackage{datatool}
\usepackage{array}
\usepackage{placeins}
\usepackage{float}
\usepackage{geometry}       
\usepackage{pifont} 
\usepackage{makecell}        
\usepackage{dblfloatfix}

\definecolor{improvement}{RGB}{144,238,144} 
\definecolor{degradation}{RGB}{255,182,193} 
\definecolor{nochange}{RGB}{255,255,255}

\newcommand{\heading}[1]{\noindent\textbf{#1}}

\usepackage[normalem]{ulem} 
\newcommand{\stkout}[1]{\ifmmode\text{\sout{\ensuremath{#1}}}\else\sout{#1}\fi}

\ifdraft

\newcommand{\added}[1]{\textcolor{blue}{#1}}
\newcommand{\deleted}[1]{\textcolor{red}{\stkout{#1}}}

\newcommand{\deletedfloat}[1]{}
\newcommand{\commented}[1]{\textcolor{blue}{#1}}
\else

\newcommand{\added}[1]{#1}
\newcommand{\deleted}[1]{}

\newcommand{\deletedfloat}[1]{}
\newcommand{\commented}[1]{}
\fi

\newcommand\citep[1]{\cite{#1}}
\newcommand\citet[1]{\cite{#1}}

\newcommand{\dbname}{\emph{MIMIC-III-Ext-PPG}}

\def\BibTeX{{\rm B\kern-.05em{\sc i\kern-.025em b}\kern-.08em
    T\kern-.1667em\lower.7ex\hbox{E}\kern-.125emX}}
\begin{document}

\title{Deriving Health Metrics from the Photoplethysmogram: Benchmarks and Insights from \dbname}

\author{\IEEEauthorblockN{Mohammad Moulaeifard\IEEEauthorrefmark{1}, Philip J. Aston\IEEEauthorrefmark{2}\IEEEauthorrefmark{3}, Peter H. Charlton\IEEEauthorrefmark{4}, Nils Strodthoff\IEEEauthorrefmark{1}*}

\IEEEauthorblockA{\IEEEauthorrefmark{1}AI4Health Department, Oldenburg University, Oldenburg, Germany}

\IEEEauthorblockA{\IEEEauthorrefmark{2}Department of Data Science and AI, National Physical Laboratory, Teddington, United Kingdom}

\IEEEauthorblockA{\IEEEauthorrefmark{3}School of Mathematics and Physics, University of Surrey, Guildford, United Kingdom}

\IEEEauthorblockA{\IEEEauthorrefmark{4}Department of Public Health and Primary Care, University of Cambridge, Cambridge, United Kingdom}

\IEEEauthorblockA{*Corresponding author: nils.strodthoff@uol.de}}

\maketitle
\begin{abstract}
Photoplethysmography (PPG) is one of the most widely captured biosignals for clinical prediction tasks, yet PPG-based algorithms are typically trained on small-scale datasets of uncertain quality, which hinders meaningful algorithm comparisons. We present a comprehensive benchmark for PPG-based clinical prediction using the \dbname~dataset, establishing baselines across the full spectrum of clinically relevant applications: multi-class heart rhythm classification, and regression of physiological parameters including respiratory rate (RR), heart rate (HR), and blood pressure (BP). Most notably, we provide the first comprehensive assessment of PPG for general arrhythmia detection beyond atrial fibrillation (AF) and atrial flutter (AFLT), with performance stratified by BP, HR, and demographic subgroups. Using established deep learning architectures, we achieved strong performance for AF detection (AUROC = 0.96) and accurate physiological parameter estimation (RR MAE: 2.97 bpm; HR MAE: 1.13 bpm; SBP/DBP MAE: 16.13/8.70 mmHg). Cross-dataset validation demonstrates excellent generalizability for AF detection (AUROC = 0.97), while clinical subgroup analysis reveals marked performance differences across subgroups by BP, HR, and demographic strata. These variations appear to reflect population-specific waveform differences rather than systematic bias in model behavior.
This framework establishes the first integrated benchmark for multi-task PPG-based clinical prediction, demonstrating that PPG signals can effectively support multiple simultaneous monitoring tasks and providing essential baselines for future algorithm development.

\end{abstract}


\section{Introduction}
\label{sec:I}


\IEEEPARstart{P}hotoplethysmography (PPG) has become one of the most widely available biosignal monitoring technologies in contemporary healthcare, delivering non-invasive, continuous physiological monitoring functionality in a wide range of clinical contexts \cite{allen2007photoplethysmography, charlton2022wearable}. PPG sensors are used from the intensive care unit (ICU) to consumer wearables, offering real-time physiological information about cardiovascular and respiratory function through the optical measurement of the blood volume variations in peripheral tissues \cite{elgendi2012standard, castaneda2018review}.

\heading{PPG-based prediction tasks} Recent machine learning advances have shown the promise of PPG for arrhythmia detection \cite{liu2022multiclass, torres2020multitask, MOULAEIFARD2026109831}, BP estimation \cite{gaurav2016cuffless, ding2017continuous, moulaeifard2025generalizable}, and RR  estimation \cite{karlen2013photoplethysmogram, addison2015pulse}, to name just a few prominent applications.

\heading{Limitations of existing benchmarking studies}
Although several benchmark studies have attempted to evaluate PPG-based prediction tasks, these efforts are hindered by certain methodological limitations. The benchmarking studies for RR estimation, e.g., \cite{charlton2016assessment, pimentel2016toward}, HR estimation, e.g., \cite{sicbaldi2025benchmarking}, and BP prediction, e.g., \cite{moulaeifard2025machine}, provide useful and informative insights. \deleted{However, these benchmarks remain limited in scope, restricting their generalizability, and do not address multi-task or metadata-rich prediction.}\added{However, these studies  mostly evaluate isolated prediction tasks and rely on relatively constrained datasets, which limits their ability to generalize across populations and physiological conditions. Also, they rarely incorporate multi-task learning or leverage rich clinical metadata, leaving important aspects of real-world physiological monitoring unexplored.} 

\added{\heading{Finegrained rhythm classification from PPG} In terms of arrhythmia classification, existing PPG-based studies have focused on addressing limited and narrowly defined scenarios, such as AF vs. non-AF, abnormal rhythm detection beyond AF, or multiclass settings with a relatively low number of rhythm classes \cite{polania2015method,han2020premature,jeanningros2024non,liu2022multiclass}.
As a result, the fine distinction between rhythms across a large number of clinically relevant classes remains largely unexplored. Moreover, the evaluation of model performance on clinically relevant classes, such as those based on BP, HR, or demographic features, is not commonly performed. All these issues underscore the need for a comprehensive and multi-task benchmarking approach to PPG analysis, as presented in this work.}


\begin{table*}[!htbp]
\centering
\renewcommand{\arraystretch}{1.2}
\caption{Demographic and clinical characteristics of the study population across different physiological subsets. Values are presented as mean $\pm$ SD or percentage (\%).}
\begin{tabular}{lcccc}
\toprule
\textbf{Metric} & \textbf{Heart rhythm} & \textbf{RR} & \textbf{HR} & \textbf{BP } \\
\midrule

\textbf{Number of subjects} & \added{6,189} & 4,695 & 5,924 & 2,391 \\
\textbf{Number of segments (30s)} & 6,399,754 & 5,546,169 & 5,187,183 & 2,468,693 \\
\textbf{Total Duration (h)} & $\sim$53,331 & $\sim$46,218 & $\sim$43,227 & $\sim$20,572 \\

\textbf{Age (years, mean $\pm$ SD)} & 64.1 $\pm$ 17.0 & 65.0 $\pm$ 16.1 & 63.9 $\pm$ 16.9 & 63.2 $\pm$ 16.2 \\

\textbf{Weight (kg, mean $\pm$ SD)} & 82.2 $\pm$ 22.6 & 82.0 $\pm$ 21.8 & 82.3 $\pm$ 22.6 & 83.2 $\pm$ 21.9 \\

\textbf{Height (cm, mean $\pm$ SD)} & 169.5 $\pm$ 10.5 & 169.7 $\pm$ 10.5 & 169.4 $\pm$ 10.5 & 169.7 $\pm$ 10.4 \\

\textbf{Gender (female, \%)} & 43.9 & 43.3 & 44.0 & 42.2 \\

\textbf{Ethnicity (\%)} & & & & \\
\textbf{- White} & 72.0 & 73.5 & 72.0 & 71.4 \\
\textbf{- Black} & 9.2 & 9.3 & 9.1 & 7.7 \\
\textbf{- Hispanic} & 4.1 & 3.8 & 4.1 & 3.8 \\
\textbf{- Asian} & 2.8 & 2.8 & 2.9 & 3.0 \\
\textbf{- Other} & 11.9 & 10.6 & 11.9 & 14.1 \\

\bottomrule
\end{tabular}
\label{tab:tab1}
\end{table*}

\heading{Dataset considered in this work}
The recent large-scale, quality-assessed PPG dataset \dbname{} \cite{PhysioNet-mimic-iii-ext-ppg-1.1.0}  addresses these limitations by offering several key advantages over existing datasets. It provides the most comprehensive task coverage, includes over 6.3 million 30-second PPG segments from 6,189 intensive care patients, supporting both extensive heart rhythm classification (26 classes) and physiological parameter regression (BP, HR, RR). The dataset includes rich clinical metadata encompassing demographics, ethnicity, biometrics, and comorbidities, enabling thorough subgroup analysis. 


\heading{Contributions} In this work, we introduce a comprehensive benchmarking framework consolidating both classification and regression tasks for PPG-based clinical prediction. Our rigorous assessment includes heart rhythm classification at multiple granularities, physiological parameter regression, such as RR, HR, and BP estimation. We set up performance baselines using well-known deep learning architectures and investigate model generalizability through out-of-distribution testing and clinical subgroup analysis.
More specifically, the contributions of this study are the following: (1) we presented a benchmarking protocol for comprehensive PPG clinical prediction, demonstrating the feasibility of multi-task physiological monitoring from PPG signals; (2) we provided essential performance benchmarks for all major PPG clinical applications, facilitating standardized assessment for prospective algorithmic advancements; (3) in particular, we demonstrated the feasibility of finegrained arrhythmia classification based on PPG data alone; and (4) we revealed task-specific performance discrepancies among clinical \added{and demographic} subgroups (e.g., varying hypertension levels, HR categories, BMI, gender, and ethnicity in heart rhythm classification), offering critical insights for real-world deployment.

\section{Materials and Methods}

\subsection{Dataset and Task Definitions}

\heading{Dataset summary} We used the \dbname~dataset \cite{PhysioNet-mimic-iii-ext-ppg-1.1.0,Goldberger2020:physionet}, a comprehensive PPG-based benchmark (Table \ref{tab:tab1}). \added{This dataset is derived from the MIMIC-III Waveform Database Matched Subset \cite{mimiciiiwaveform}, containing recordings from critically-ill patients. It comprises 6,399,754 30-second PPG segments from 6,189 subjects, all annotated for heart rhythm classification. In addition, subsets provide labels for RR (4,695 subjects, 5,546,169 segments), HR (5,924 subjects, 5,187,183 segments), and BP (2,391 subjects, 2,468,693 segments).}

\heading{Heart rhythm classification tasks} We trained and evaluated models across two primary categories (regression and classification tasks) encompassing the full spectrum of clinically relevant PPG applications. Table \ref{tab:heart-rhythms_after_filter} presents each heart rhythm, the associated number of unique patients, the number of 30-second samples, and the corresponding rhythm abbreviation used in this paper. For heart rhythm classification, we distinguished three scenarios (tasks):
\begin{itemize}
\item \emph{Robust AF/AFLT detection (AF task)}: \added{A multi-label classification task with two binary output nodes—one for AF detection and one for AFLT detection. This task setup aligns with typical AF/AFLT detection settings considered in the literature.} 

\item \emph{Sinus versus atrial arrhythmia discrimination (SAA task)}: \added{A multi-label binary classification task with two binary output nodes -- one to detect sinus rhythm (SINUS) and one to detect atrial arrhythmias (AF+AFLT).} SINUS comprises SR, STACH, SBRAD, and SARRH, while AF+AFLT includes AF and AFLT. This task assesses the model's ability to detect atrial rhythm disturbances while remaining robust to physiological variations within sinus rhythm.

\item \emph{Comprehensive rhythm classification (ARRH task)}: \added{A multi-label classification task with 13 binary output nodes, each corresponding to a specific rhythm type. We included all 13 rhythm classes with more than 1,000 30-second samples, while all other well-defined rhythm types are included in OTHER. This task evaluates the possibility of fine-grained rhythm classification based solely on PPG signals.}
\end{itemize}

\added{It is worth noting that in all scenarios,  we excluded all atrioventricular blocks as well as left and right bundle branch blocks (LBBB and RBBB), as they can co-occur with different rhythm types. 
For all three tasks, we include all samples from the respective heterogeneous OTHER rhythm sample group with zero labels for all outputs to improve robustness to non-target rhythms. The complete mapping table is given Table~\ref{tab:heart-rhythms_after_filter}.}

\heading{Regression tasks}
For regression tasks, we evaluated continuous prediction of RR and HR (breaths/beats per minute), and simultaneous SBP/DBP estimation (mmHg) from the PPG. All labels are provided as columns in the \dbname{} dataset.

\subsection{Model Architectures and Experimental Setup}

\heading{Model} We used two established deep learning architectures: XResNet1d101 \cite{Strodthoff:2020Deep}, a state-of-the-art ResNet architecture, and LeNet1D \cite{Wagner2024}, a shallow convolutional model adapted to 1D time series  as a baseline. Both models showed robust performance in a recent benchmarking study \cite{moulaeifard2025machine}. The XResNet1d101 model was chosen as a representative for a comparably deep, parameter-heavy model, and LeNet1D as a representative for a lightweight model.

\heading{Experimental setup} 
\added{
\dbname{} provided raw PPG segments without flatlines or NaN values. Further pre-processing of the signals was performed by utilizing the \texttt{ppg\_clean} function provided by the NeuroKit2 library \cite{makowski2021neurokit2}. The \texttt{Elgendi} algorithm was utilized, which includes a Butterworth bandpass filter between 0.5 and 8~Hz to reduce the baseline drift and high-frequency noise. The sampling rate was set at 125~Hz, and the models were trained at full input resolution of each segment, which was equivalent to 3750 time steps (30 seconds).}

We leveraged the provided 10 folds, stratified according to heart rhythm events, age distributions, and cardiovascular diagnoses. The first seven folds were used for training, the eighth fold for validation, and the ninth and tenth folds for testing. For each experiment, an effective batch size of 512 was used through gradient accumulation. The learning rates were set to 0.001 for all models, and models were trained for 50 epochs. The training was implemented using the AdamW optimizer \cite{loshchilov2017decoupled}. As a simple measure to reduce overfitting, we performed model selection based on the validation set score, i.e., during training, we kept track of the validation set score and selected the model with the best validation set score for evaluation on the test set. We used the macro averaged area under the receiver operating curve (AUROC) as a threshold-independent metric to characterize the overall discriminative power of the model for classification tasks and the mean absolute error (MAE) as the main metric for regression tasks.

\heading{BP estimation} For BP estimation, consistent with previous studies, e.g., \cite{xiao2024advancing,Moulaeifard2025}, we employed two output nodes to jointly predict SBP and DBP, leveraging possibly shared physiological features to enhance model performance for both SBP and DBP.

\heading{Arrhythmia classification} In all arrhythmia detection cases, we handled the heterogeneous set of remaining rhythm types by using a one-hot encoding and applying binary cross-entropy across all output channels. Although each sample belongs to exactly one rhythm class, this formulation treats the problem as a multi-class prediction with independent sigmoid outputs rather than a softmax layer, see \cite{wightman2021resnet} for similar approaches in computer vision. Consequently, this setup intentionally does not explicitly model a separate “remaining” class OTHER; therefore, the corresponding results table does not include a row for the class OTHER.

\heading{External validation}
Model generalizability was assessed through out-of-distribution evaluation by testing on the Liu et al. dataset \cite{liu2022multiclass} as an external validation dataset. The external dataset contains 6 heart rhythm types, namely AF, SR, premature ventricular contraction, premature atrial contraction, VTACH, and SVTACH, which we mapped to our classification scenarios, see Table \ref{tab:liu_mapping} in the supplementary material. It is worth noting that the external dataset is tested using the full input resolution of 1000 time steps (10 seconds). The fact that all considered model architectures involve a global average pooling layer allows for flexible input lengths at inference time, while still leveraging the model trained on \dbname{}.

\heading{Subgroup analyses}
One of the strengths of the \dbname{} dataset is the simultaneous availability of heart rhythm labels alongside other physiological parameters, including HR, BP, and demographic information. We used this information to stratify the arrhythmia detection performance by:

\begin{enumerate}
    \item \added{\textbf{BP categories} according to standard clinical guidelines 
    \cite{flack2020blood, sharma2023hypotension}, see Table~\ref{tab:bp_classification} in the supplementary material.}

    \item \added{\textbf{HR categories} based on \cite{AHA_2005_brady_tachy}, 
    see Table~\ref{tab:hr_classification} in the supplementary material.}

    \added{\item Demographic/biometric categories:}
    \begin{enumerate}
        \item \added{\textbf{BMI categories} based on World Health Organization (WHO) recommendations 
        \cite{who2000obesity}, see Table~\ref{tab:bmi_categories}.}

        \item \added{\textbf{Ethnicity categories} were aggregated into five groups 
        (White, Black, Asian, Hispanic, and Other), aligned with 
        U.S. Office of Management and Budget Statistical Policy Directive 
        No.~15 (1997) \cite{office1997revisions} and consistent with prior machine learning 
        studies using MIMIC-III (e.g., \cite{chen2020exploring}), see Table~\ref{tab:ethnicity_mapping}.}

        \item \added{\textbf{Gender categories} as recorded in the MIMIC-III database 
        \cite{johnson2016mimic}, categorized as male and female according to the 
        structured electronic health record fields.}
    \end{enumerate}
    
\end{enumerate}

\heading{Performance evaluation and metrics}

\added{Regression tasks are evaluated using the Mean Absolute Error (MAE). We also assess the agreement between predictions and reference values in more detail. Bland-Altman analyses were also performed \cite{bland1986statistical}. We determined bias as well as the corresponding limits of agreement (LoA).}

\added{
Rhythm classification performance is assessed using the area under the receiver operator characteristic curve (AUROC). To further investigate the performance of the models, we also report sensitivity while fixing the decision threshold such that we reach a specificity $\geq 0.8$, and report specificity while fixing the threshold such that sensitivity $\geq 0.80$.}

\begin{table*}[ht]
\centering
\caption{In-distribution classification performance for robust AF classification (AF task), sinus versus atrial arrhythmia discrimination (SAA task), and general arrhythmia classification (ARRH task) in terms of label-dependent AUROC for the two models considered in this work.} 
\renewcommand{\arraystretch}{1.4}

\resizebox{\textwidth}{!}{
\begin{tabular}{llrrrrrr}
\toprule
& & \multicolumn{2}{c}{\textbf{AUROC}} 
& \multicolumn{2}{c}{\textbf{Sensitivity (Specificity $>$ 0.8)}} 
& \multicolumn{2}{c}{\textbf{Specificity (Sensitivity $>$ 0.8)}} \\
\cmidrule(lr){3-4}
\cmidrule(lr){5-6}
\cmidrule(lr){7-8}

\textbf{Task} & \textbf{Rhythm} 
& \textbf{XResNet1d101} & \textbf{LeNet1D}
& \textbf{XResNet1d101} & \textbf{LeNet1D}
& \textbf{XResNet1d101} & \textbf{LeNet1D} \\
\midrule

\multirow{2}{*}{\rotatebox{90}{AF}}
& AF 
& 0.96 & 0.94 
& 0.92 & 0.92 
& 0.98 & 0.96 \\

& AFLT 
& 0.77 & 0.78 
& 0.62 & 0.64 
& 0.61 & 0.62 \\

\midrule

\multirow{2}{*}{\rotatebox{90}{SAA}}

& SINUS 
& 0.92 & 0.90 
& 0.89 & 0.86 
& 0.86 & 0.84 \\

& AF+AFLT 
& 0.95 & 0.94 
& 0.93 & 0.92 
& 0.98 & 0.96 \\

\midrule

\multirow{13}{*}{\rotatebox{90}{ARRH}}

& SR 
& 0.93 & 0.93 
& 0.93 & 0.92 
& 0.89 & 0.89 \\

& STACH 
& 0.97 & 0.97
& 0.97 & 0.97 
& 0.97 & 0.97 \\

& AF 
&0.96 & 0.94 
& 0.94 & 0.92 
& 0.96 & 0.96 \\

& SBRAD 
& 0.96 & 0.96 
& 0.95 & 0.95 
& 0.98 & 0.98 \\

& VPACE 
& 0.79 & 0.80 
& 0.62 & 0.64 
& 0.60 & 0.63 \\

& AVPACE 
& 0.89 & 0.87 
& 0.81 & 0.76 
& 0.81 & 0.77 \\

& AFLT 
& 0.78 & 0.78 
& 0.65 & 0.64 
& 0.60 & 0.60 \\

& APACE 
& 0.80 & 0.78 
& 0.63 & 0.60 
& 0.61 & 0.57 \\

& SARRH 
& 0.61 & 0.66
& 0.40 & 0.49 
& 0.26 & 0.27 \\

& JR 
& 0.62 & 0.83
& 0.45 & 0.71 
& 0.20 & 0.70 \\

& SVTACH 
& 0.78 & 0.68 
& 0.65 & 0.60 
& 0.54 & 0.57 \\

& MATACH 
& 0.84 & 0.78 
& 0.72 & 0.70 
& 0.72 & 0.60 \\

& VTACH 
& 0.75 & 0.63 
& 0.60 & 0.33 
& 0.49 & 0.40 \\

\bottomrule
\end{tabular}
}

\label{tab:aftasks_results}
\end{table*}

\begin{table*}[!h]
\centering
\caption{\added{Regression performance including MAE, Bland--Altman bias, and 95\% limits of agreement (LoA) for BP, HR, and RR obtained with a median baseline predictor, XResNet1d101, and LeNet1D.}} 
\renewcommand{\arraystretch}{1.4}
\scalebox{1.05}{
\begin{tabular}{l|c|ccc|ccc}
\toprule
& \textbf{Baseline} 
& \multicolumn{3}{c}{\textbf{XResNet1d101}} 
& \multicolumn{3}{c}{\textbf{LeNet1D}} \\
\cmidrule(l){3-5}
\cmidrule(r){6-8}
\textbf{Parameter} 
& \textbf{MAE}
& \textbf{MAE} & \textbf{Bias} & \textbf{95\% LoA}
& \textbf{MAE} & \textbf{Bias} & \textbf{95\% LoA} \\
\midrule

SBP [mmHg] 
& 19.06
& 16.13 & -2.58 & [-43.16, 38.00] 
& 16.65 & -1.24 & [-43.00, 40.52] \\

DBP [mmHg] 
& 10.09
& 8.70 & 0.79 & [-21.34, 22.92] 
& 9.39 & 2.53 & [-20.38, 25.45] \\

HR [bpm] 
& 14.37
& 1.13 & -0.14 & [-8.00, 7.72] 
& 1.61 & 0.03 & [-8.77, 8.84] \\

RR [bpm] 
& 5.34
& 2.97 & -0.17 & [-10.07, 9.72] 
& 4.53 & 0.31 & [-11.89, 12.50] \\

\bottomrule
\end{tabular}
}
\label{tab:regression_ba_results}
\end{table*}

\section{Results}

\subsection{Arrhythmia Classification Performance}
\added{
As indicated in Table \ref{tab:aftasks_results},  the performance of AF, SAA, and ARRH tasks for in-distribution classification is provided.
In the AF task, AF detection achieved high performance for both models, with AUROC values of 0.96 for XResNet1d101 and 0.94 for LeNet1D, which translates into high sensitivity (0.92 for both models) and high specificity (0.98 vs.\ 0.96) at operating points of specificity/sensitivity $\geq$0.80, respectively. AFLT detection showed lower performance, with AUROC values around 0.77–0.78, sensitivity 0.62–0.64, and specificity 0.61–0.62.
In SAA, SINUS classification showed high performance for both models, where AUROC for XResNet1d101 and LeNet1D were 0.92 and 0.90, respectively, and correspondingly high values of sensitivity and specificity, where sensitivity is equal to 0.89 for XResNet1d101 and 0.86 for LeNet1D, and specificity is equal to 0.86 for XResNet1d101 and 0.84 for LeNet1D. 
The performance of the AF+AFLT detection remained high for both models with AUROC values of 0.95 and 0.94, sensitivity of 0.93 and 0.92, and specificity of 0.98 and 0.96 for XResNet1d101 and LeNet1D, respectively.
Within the ARRH task, several rhythms showed high performance for both models, including STACH (AUROC = 0.97), SR (AUROC = 0.93), AF (AUROC = 0.94-0.96), and SBRAD (AUROC = 0.96), along with high values of sensitivity and specificity. However, some of the less common rhythms showed less performance and variability. For instance, VPACE achieved AUROC values of 0.79-0.80, compared to SARRH and JR, which showed lower AUROC and threshold metrics. Interestingly, LeNet1D showed higher performance for JR (AUROC = 0.83, sensitivity = 0.71, specificity = 0.70) compared to XResNet1d101. Overall, it is evident that XResNet1d101 showed slightly higher performance for most of the rhythms, although both models demonstrated comparable results for several classes.}

\begin{figure*}[htbp]
    \centering
    \includegraphics[scale=0.4, keepaspectratio]{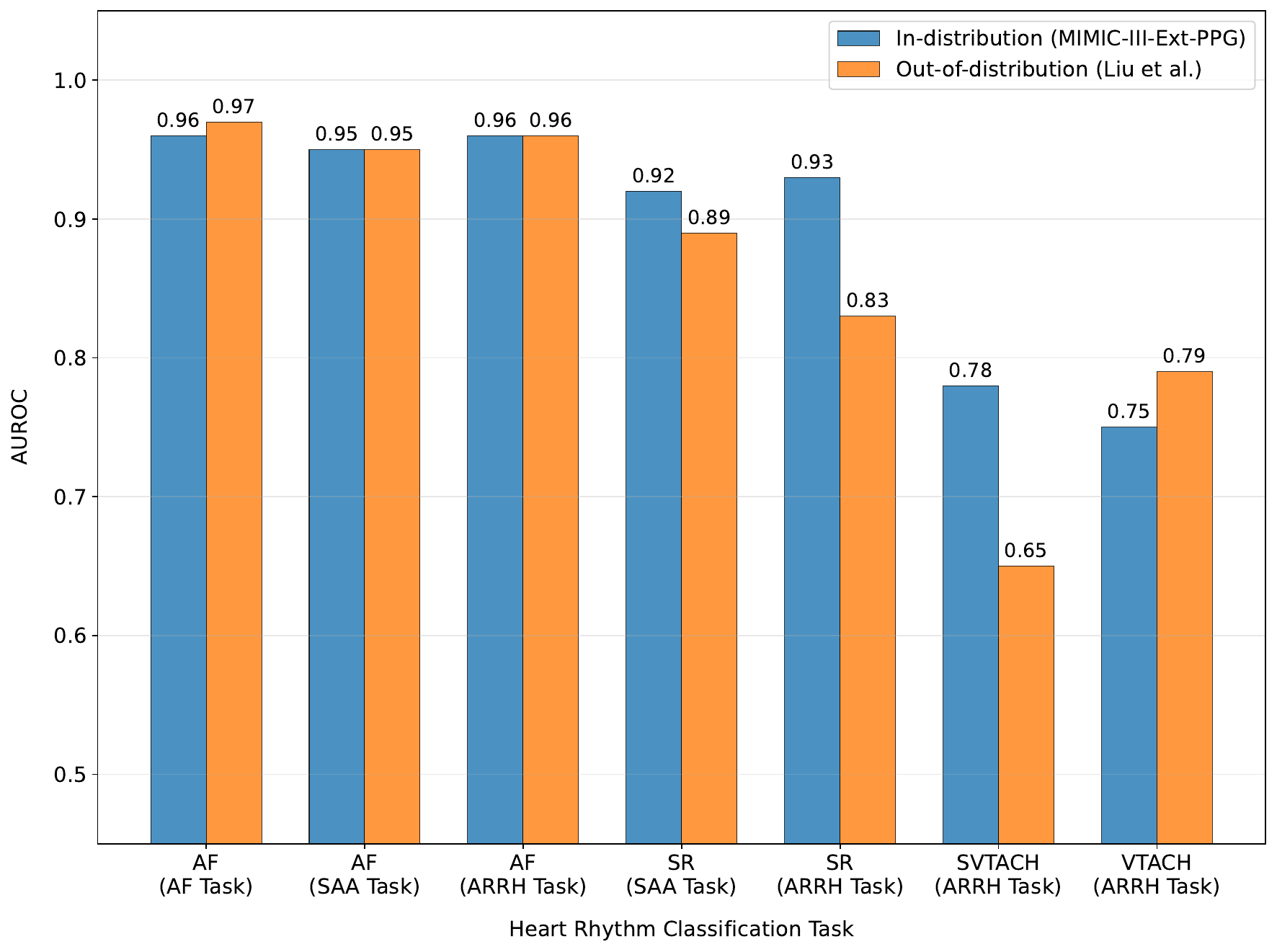}
    \caption{\added{External validation of the rhythm classification performance (using XResNet1d101)} demonstrating robust generalization of PPG-based rhythm classification, with AF detection showing exceptional cross-dataset transferability.}
    \label{fig:ppg_generalization}
\end{figure*}

\subsection{Regression Task Performance}
Table \ref{tab:regression_ba_results} presents the performance on all regression tasks. XResNet1d101 achieved superior performance across RR and HR prediction tasks: RR prediction with an MAE of 2.97 bpm compared to LeNet1D's 4.53 bpm, and HR estimation with an MAE of 1.13 bpm versus 1.61 bpm for LeNet1D. In terms of BP estimation, XResNet1d101 demonstrated slightly better performance with MAEs of 16.13/8.70 mmHg (SBP/DBP) compared to LeNet1D's 16.65/9.39 mmHg.


\added{
Furthermore, the comparative Bland–Altman analysis (model bias and LoA) for SBP, DBP, HR, and RR shows that both models have near-zero bias for all parameters. The best performance is observed for HR, where the LoAs are relatively narrow (approximately $\pm$8\,bpm) compared with the physiological variability in the dataset (median 86.2\,bpm, IQR 75.0–98.7\,bpm). Performance is poorer for RR where the LoAs are wider in comparison to the physiological variability (LoAs of approximately $\pm$10\,bpm compared to median 19.9\,bpm, IQR 15.7–24.8\,bpm). Performance was worst for SBP, with LoAs of approximately $\pm$40\,mmHg, which is considerably larger than the interquartile range of SBP values in the dataset (median 121.3\,mmHg, IQR 105.6–139.6\,mmHg). These findings highlight that while HR estimation from PPG is highly reliable, RR estimation remains moderately challenging, and cuffless BP estimation based on PPG alone using these techniques is currently unreliable.}

\subsection{Out-of-Distribution Generalization}
Figure \ref{fig:ppg_generalization} shows the out-of-distribution evaluation results for the XResNet1d101 model trained on \dbname~and tested on the external dataset by Liu et al\cite{liu2022multiclass,bashar2020atrial}. It is worth recalling that, since the external dataset lacks several rhythm classes found in \dbname, our results are limited to the shared classes such as AF, SR, SVTACH, and VTACH.

\added{The AF task scenario demonstrated exceptional cross-dataset transferability with an AUROC of 0.97 for AF detection, in line with the strong in-distribution performance (AUROC 0.96) and confirming the robustness of PPG-based AF detection algorithms across different patient populations and recording conditions. Similarly, the SAA task demonstrated strong generalization for atrial arrhythmia detection (AUROC of 0.95 vs. internal AUROC 0.95) and slightly worsened discrimination for SR detection (AUROC of 0.89 vs. internal AUROC 0.92) for the external dataset.
For the arrhythmia classification (ARRH task), the XResNet1d101 model retains high performance for AF detection (AUROC 0.96 vs. internal AUROC 0.96). However, a notable degradation in performance is seen for SR classification (AUROC 0.83 vs. internal AUROC 0.93). A slight improvement in performance is seen for VTACH (AUROC 0.79 vs. internal AUROC 0.75), but for SVTACH (AUROC 0.65 vs. internal AUROC 0.78), the model performs significantly worse on the external dataset.}
\added{One possible explanation for the drop in SR detection performance in the ARRH scenario could be the fact that the model predicts four finegrained SR subclasses, but is evaluated only on a coarse SR class in the external dataset. This result is in line with the strong external validation result for the coarse SR class (AUROC 0.89) for the SAA task.}

\subsection{Subgroup Analyses}
\added{In this section, we analyzed the arrhythmia classification performance (all three tasks) of the XResNet1d101 model stratified by BP, HR, gender, BMI, and ethnicity categories (when available).}

\added{
\heading{Stratification according to BP} Tables \ref{tab:bp_hr_auroc} and \ref{tab:bp_hr_threshold} summarize rhythm classification performance across BP categories. In the AF task, the performance of the AF detection method is consistently high across BP groups with AUROC values between 0.95 and 0.99, coupled with high sensitivity (0.92–0.99) and specificity (0.98–1.00). In the case of AFLT detection, the performance is lower and shows moderate variability across BP groups, with AUROC values ranging from 0.77 to 0.84 and corresponding sensitivity and specificity values between 0.54–0.78 and 0.65–0.79, respectively.
In the SAA task, SINUS classification achieved AUROC values between 0.88 and 0.97, while AF+AFLT detection was high and stable across BP groups (0.96–0.99) with high sensitivity and specificity.
For the ARRH task, common rhythms such as SR (0.92-0.96), STACH (0.96-0.98), and AF (0.95-0.99) had a stable performance across all BP groups. On the other hand, several less common rhythms had a less stable performance across the BP groups, such as VPACE (0.65–0.91), SVTACH (0.47–0.93), and VTACH (0.42–0.92), which was also reflected in more variable sensitivity and specificity values.}

\added{
\heading{Stratification according to HR}
Tables \ref{tab:bp_hr_auroc} and \ref{tab:bp_hr_threshold} present the performance of the rhythm classification algorithms across HR categories. In the AF task, the performance of AF detection increased with the increase of HR.The AUROC of AF detection increased from 0.88 in the bradycardia category to 0.97 in the tachycardia category. Meanwhile, the sensitivity and specificity of AF detection increased from 0.79 / 0.78 to 0.96 / 0.97. In the AFLT detection task, the AUROC of AFLT detection increased from 0.74 to 0.86. Meanwhile, the sensitivity and specificity of AFLT detection increased from 0.51 / 0.50 to 0.73 / 0.76.
In the SAA task, SINUS classification improved from 0.84 AUROC in bradycardia to 0.96 in tachycardia, with corresponding increases in sensitivity and specificity (0.70 / 0.74 vs.\ 0.97 / 0.95), while AF+AFLT detection remained consistently strong (0.87–0.97 AUROC).
Within the ARRH task, several rhythms showed reduced performance in bradycardia, such as SR (0.72 AUROC, 0.51 / 0.49 sensitivity/specificity). In contrast, some rhythms showed improved performance at higher HR values, including APACE (0.96 AUROC, 0.97 / 0.96 sensitivity/specificity).}

\added{ 
\heading{Stratification according to gender}
Table~\ref{tab:gender_bmi_ethnicity_auroc} shows AUROC scores stratified according to gender, BMI and ethnicity, see Table~\ref{tab:aftasks_demographics_threshold} for corresponding sensitivity/specificity metrics.
Across gender groups, rhythm classification performance remained largely comparable. In the AF task, AF detection achieved similar AUROC values in females and males (0.95 vs.\ 0.96) with nearly identical sensitivity and specificity (0.92/0.98 vs.\ 0.93/0.98). AFLT detection showed moderate variation (0.84 vs.\ 0.75 AUROC, 0.74/0.74 vs.\ 0.60/0.59 sensitivity/specificity). In the SAA task, both SINUS (0.91 vs.\ 0.93) and AF+AFLT (0.95 vs.\ 0.95) detection remained consistent across genders with comparable threshold metrics. Within the ARRH task, most common rhythms also showed similar performance (e.g., SR 0.93 vs.\ 0.92, STACH 0.97 vs.\ 0.97), although some rhythms exhibited moderate variation such as VPACE (0.71 vs.\ 0.84 AUROC).}

\added{ 
\heading{Stratification according to BMI}
Moderate differences were observed across BMI groups. In the AF task, AF detection increased from 0.92 AUROC in normal BMI to 0.98 in obese subjects, with sensitivity/specificity increasing from 0.89/0.94 to 0.97/0.99. AFLT detection remained relatively stable (0.72–0.75 AUROC). In the SAA task, SINUS classification improved from 0.84 (normal BMI) to 0.96 (obese) with corresponding increases in sensitivity and specificity (0.69/0.72 vs.\ 0.97/0.94), while AF+AFLT detection remained high across BMI groups (0.93–0.97 AUROC). Similar trends were observed for several ARRH rhythms such as SR (0.87–0.95 AUROC) and VPACE (0.65–0.95 AUROC).}

\added{ 
\heading{Stratification according to ethnicity}
For all ethnicity groups, most of the rhythms demonstrated comparable performance. In the AF task, the AUROC of AF detection varied from 0.95 to 0.98, with high sensitivity and specificity (e.g., 0.90/0.97 in the White ethnicity, 0.96/0.99 in the Asian ethnicity). In the SAA task, the AUROC of SINUS classification varied from 0.82 to 0.98, while the AUROC of AF+AFLT detection was stable at 0.95 to 0.98. Within the ARRH task, common rhythms such as SR (0.91–0.98) and STACH (0.97–0.99) showed stable performance across ethnic groups, while some less frequent rhythms showed larger variability.}

\section{Discussion}

\heading{Heart Rhythm Classification Performance}
Our comprehensive benchmarking establishes the performance baselines for PPG-based clinical prediction tasks on the recently published \dbname{} dataset. The strong performance in AF detection (AUROC $\approx 0.95$) confirms the clinical utility of the PPG for the most common arrhythmia detection application in a large and diverse ICU population. Similarly, the SAA task achieved consistently high in-distribution performance (AUROC $\sim$ 0.92 – 0.95), supporting reliable discrimination between sinus rhythms and atrial arrhythmias using PPG alone. Compared to other state-of-the-art AF classification studies, the advantages of leveraging our large, diverse dataset with high AUROC, even in the challenging ICU setting, are clear. While high AUROC ($\sim 0.99\text{--}1.0$) is often achieved on limited cohorts (e.g., $N < 50$ in \cite{aliamiri2018recurrent, gotlibovych2018cnn, nemati2016robust}) or ambulatory data, performance often drops significantly when scaling up (e.g., AUROC $0.72$ in \cite{tison2018passivenejm}). Furthermore, \cite{moulaeifard2025machine} on the DeepBeat dataset \cite{torres2020multi} reported AUROC in the $\sim 0.8\text{--}0.9$ range with less than 200 patients. Our results (AUROC $\approx 0.95$) on a large and diverse clinical population (up to 6,189 patients, 53,331 hours of data) demonstrate superior robustness in a real-world ICU environment.

Most notably, this is the first work that systematically explores the prospects of PPG for general arrhythmia detection, demonstrating the applicability of PPG-based heart rhythm analysis beyond AF/AFLT. We believe that the currently moderate overall performance highlights the inherent challenges in extracting complex electrophysiological information from peripheral pulse morphology alone and should be seen as a challenging benchmark for the community. We also deliberately left the small classes, e.g., VTACH or JR, in the dataset, for which classification performance remained limited, to encourage future methodological advances targeting these clinically important but difficult arrhythmias.

\heading{Regression performance}
The regression task results give valuable insights into PPG's quantitative monitoring potential. 
HR estimation accuracy (MAE 1.13-1.61 bpm) outperforms recent state-of-the-art approaches, including Q-PPG's 4.41 bpm \cite{burrello2021qppg} and EnhancePPG's 3.54 bpm \cite{enhanceppg2024}, and also demonstrates superior performance compared to methods achieving 1.86-3.78 bpm under motion artifact conditions \cite{thakur2023precision}.
The RR prediction model demonstrated an MAE of 2.97-4.53 bpm, which is competitive with recent approaches, including deep learning models reporting 2.0-3.13 bpm \cite{chin2024novel, yu2023evaluation}, indicating performance characteristics that warrant investigation for ICU monitoring applications.
\added{For BP estimation, our models achieved MAE of 16.13-16.65 mmHg for SBP and 8.70-9.39 mmHg for DBP (Table \ref{tab:bp_comparison}). While our DBP performance is comparable to other MIMIC-III subset benchmarks (e.g., \cite{moulaeifard2025generalizable, sanches2024mimic} reported $\sim 9\text{--}10\ \si{\mmHg}$), our SBP MAE falls within, but toward the upper bound of their reported range ($\sim 13\text{--}17\ \si{\mmHg}$). Finally, all model comparisons that do not rely on an identical dataset and splits have to be taken with a grain of salt.}
Furthermore, Table \ref{tab:regression_ba_results} reveals substantial variability in BP estimation, with particularly wide LoA ranges for SBP and DBP, indicating large per-sample fluctuations despite low average error. This trend is also observed in \cite{moulaeifard2025generalizable}, which used the ICU PPG data for BP prediction.
The main conclusion on BP estimation seems to be that a larger dataset does not lead to a substantial reduction of BP prediction errors. Therefore, BP estimation from PPG alone remains a challenging use case.

\heading{Architecture Comparison and Model Selection}
The comparable performance between XResNet1d101 and LeNet1D across most tasks indicates that PPG signal features may not require the depth in architecture that is commonly necessary for image recognition tasks. While XResNet1d101 achieved slightly higher performance in AF detection, LeNet1D demonstrated competitive performance with a substantially simpler architecture, and showed slightly superior performance in BP estimation, indicating that simpler models may actually be advantageous for certain PPG applications, potentially offering better generalization and computational efficiency \cite{moulaeifard2025machine}. The minimal performance differences between architectures also indicate that data quality and preprocessing may be more critical factors than model complexity for PPG applications.

\heading{Out-of-Distribution Generalization}
\added{Overall, the external validation results indicate excellent generalization capabilities of the rhythm classification models, showing good agreement between the performance levels achieved on internal and external datasets. The most striking deterioration in external vs. internal model performance occurred for supraventricular tachycardia, where already the internal model performance is not outstanding. As SVTACH is one of the least populated rhythm types, we see it as very likely that the model performance could improve with more training data.}


\added{
\heading{Stratification according to BP}
The BP stratification analysis indicates that rhythm classification performance remains largely stable across blood pressure categories for the principal rhythms. AF detection maintained consistently high. Similar robustness was observed for AF+AFLT detection in the SAA task, suggesting that the learned representations for atrial rhythm irregularity are largely preserved across different hemodynamic states. 
In contrast, several less frequent arrhythmias showed greater variability across BP strata. Rhythms such as VPACE, SVTACH, and VTACH exhibited larger fluctuations in AUROC as well as sensitivity and specificity values across BP groups. This variability may reflect both the smaller number of available samples for these rhythms and the physiological effects of blood pressure on peripheral pulse characteristics. Variations in vascular tone and arterial stiffness associated with different blood pressure states are known to alter the morphology and timing characteristics of the PPG waveform \cite{allen2007photoplethysmography}, which may contribute to increased classification variability for rhythm types with more subtle waveform patterns. Overall, these results suggest that BP variations have a limited influence on the detection of dominant rhythms such as AF and SR, while rarer arrhythmias may be more sensitive to physiological variability and dataset imbalance.}

\added{
\heading{Stratification according to HR}
The HR subgroup analysis showed more differentiated performance differences than those observed across BP categories. The performance in AF detection improved with increasing heart rate (AUROC increasing from 0.88 in bradycardia to 0.97 in tachycardia). These improvements in performance were also found for AFLT and SINUS detection, indicating that higher heart rates facilitate clearer discrimination of atrial rhythm irregularities.
However, several rhythms showed reduced performance in bradycardic segments. For instance, SR classification performance decreased in bradycardia, and for STACH, specificity was significantly decreased in bradycardia. These findings imply that lower heart rates may reduce the distinctiveness of temporal rhythm patterns in PPG signals.
These findings are consistent with prior studies showing that heart rate strongly affects PPG waveform morphology and rhythm detectability. AF detection algorithms commonly rely on irregular pulse-to-pulse intervals derived from PPG signals. In AF, these intervals become highly variable compared to the regular spacing observed in SR, providing an important cue for rhythm classification \cite{pereira2020photoplethysmography}. In summary, the proposed HR stratification results emphasize the importance of the heart rate as a significant factor for PPG waveform rhythm classification performance.}

\added{
\heading{Stratification according to gender}
The gender-based analysis revealed largely comparable rhythm classification performance between female and male subjects, which is an important indicator of algorithmic fairness across sex groups. AF detection achieved similar AUROC values across genders (0.95 vs.\ 0.96) with nearly identical sensitivity and specificity, suggesting that the model generalizes well across sex-related physiological differences. Moderate variations were observed for certain rhythms such as AFLT and VPACE. These differences may reflect known sex-related variations in cardiovascular physiology and peripheral pulse morphology, which can influence PPG waveform characteristics \cite{dehghanojamahalleh2019sex}. Overall, the small differences observed between genders indicate that the proposed model maintains fair and robust rhythm classification performance across sexes, consistent with recommendations for equitable medical AI systems \cite{seyyed2021underdiagnosis}}

\added{
\heading{Stratification according to BMI}
The BMI-based analysis revealed moderate performance differences across body composition groups. In particular, AF and SINUS detection showed improved performance in obese subjects compared with normal BMI groups. These findings may reflect the influence of body composition on peripheral pulse characteristics and vascular dynamics. Previous studies have shown that obesity can alter arterial stiffness and pulse wave propagation dynamics \cite{Nordstrand2011,Petersen2015}, which may influence peripheral pulse characteristics and consequently the discriminative features learned from PPG signals. Nevertheless, the overall classification performance remained high across BMI categories, indicating robust model generalization.}

\added{
\heading{Stratification according to ethnicity}
The ethnicity-based analysis showed broadly consistent rhythm classification performance across ethnic groups. For example, AF detection remained stable across White, Black, Asian, and Hispanic groups (AUROC 0.95–0.98) with similarly high sensitivity and specificity. These findings suggest that the learned rhythm representations generalize well across demographic groups in the dataset.}
This observation is consistent with the results in \cite{charlton2022detecting}, which did not observe significant differences in PPG beat detection on the MIMIC dataset between Black and White subjects. However, it is in stark contrast to \cite{sjoding2020racial}, where pulse oximetry was found to be less accurate in Black compared to White subjects. This could be explained by the analyses in the present study and Charlton et al \cite{charlton2022detecting} using single-channel PPG, whereas pulse oximetry \cite{sjoding2020racial} uses dual-channel PPG, potentially making it more susceptible to racial bias. In addition, taking into consideration these interactions, model calibration or cost-sensitive training could potentially help in reducing deployment biases.

\heading{Limitations and Future Directions}
Several limitations warrant consideration. First, our observation is limited to ICU patients, whose pathophysiology may limit generalizability to ambulatory or healthy populations. Also, the use of the 30-second segment length, while suitable for rhythm classification, may be insufficient for detecting paroxysmal arrhythmias or capturing long-term hemodynamic trends. Furthermore, our work is based solely on the PPG signal, which is restrictive from the application point of view, as motion artifacts cause severe signal distortion in real-world cases.

Future research should prioritize several key areas: (1) development of BP-adaptive algorithms to address the performance variability under different hemodynamic states; (2) multi-modal fusion strategies combining PPG with other readily available signals and/or static patient metadata; (3) prospectively conducted clinical trials for assessment of efficacy and safety for deployment in the field; and (4) integration of simultaneous accelerometry signals and advanced techniques, including the TROIKA algorithm \cite{zhang2014troika}, to detect heart rates and rhythms under motion conditions.

\section{Summary and Conclusion}
This study establishes a benchmarking framework for PPG-based clinical prediction, providing comprehensive performance baselines across both classification and regression tasks using the \dbname{} dataset. Our rigorous evaluation of 6.3 million PPG segments in 6,189 ICU patients illustrates the possibility of multi-task physiological monitoring by a single biosignal. \added{Regression performance revealed strong results for respiratory rate estimation, but in particular heart rate estimation and confirmed the challenging case of blood pressure prediction from PPG alone.} \added{For arrhythmia detection, we demonstrated strong performance on different application scenarios such as atrial fibrillation detection (AF) and sinus versus atrial arrhythmia discrimination (SAA). Most notably, however, this work provides the first comprehensive assessment of PPG-based general arrhythmia classification (ARRH) across 13 finegrained classes. All models achieved strong performance for common rhythms such as AF, SR, STACH, and SBRAD, with AUROC values generally above 0.93 and consistently high sensitivity and specificity with robust performance in external validation and in subgroup analyses. We hope that the presented analysis can serve as a starting point and inspiration for more detailed assessments of PPG-based prediction problems.}

\added{Overall, these results provide a comprehensive reference baseline for PPG-based physiological monitoring and demonstrate the potential of deep learning models to extract a range of clinically relevant information from a single biosignal.} Future research should prioritize BP-adaptive algorithms, multi-modal fusion approaches, and prospective clinical validation studies. The established baselines provide a foundation for regulatory approval pathways and clinical integration strategies.

\section*{Code Availability}
This work is exclusively based on publicly available datasets. The datasets underlying this study are publicly available \cite{PhysioNet-mimic-iii-ext-ppg-1.1.0,liu2022multiclass}. The source code underlying our investigations is available at  \url{https://github.com/AI4HealthUOL/MIMIC-III-Ext-PPG_benchmarking}.

\section*{Acknowledgments}
The project (22HLT01 QUMPHY) has received funding from the European Partnership on Metrology, co-financed from the European Union’s Horizon Europe Research and Innovation Programme and by the Participating States. Funding for NPL and the University of Cambridge was provided by Innovate UK under the Horizon Europe Guarantee Extension, grant numbers 10084125 and 10091955. PHC acknowledges funding from the British Heart Foundation (BHF) grant [FS/20/20/34626]. 

\section*{Competing interests}
\noindent \added{PHC performs consultancy work for wearable manufacturers who use PPG sensors in their products; however, none of these companies had any influence on the conception or realization of this study.} The other authors declare no competing interests.

\bibliographystyle{IEEEtran}
\bibliography{bibfile}

\clearpage
\appendices
\onecolumn
\section{Supplementary Tables}

\section*{Appendix}

\added{This appendix provides supplementary information related to the dataset construction, clinical categorization criteria, external dataset mappings, and detailed subgroup performance analyses.}

\begin{itemize}

\item Table~\ref{tab:heart-rhythms_after_filter} summarizes the mapping of heart rhythm annotations after applying patient and sample threshold rules, as well as their assignment across the three classification tasks used in this study.

\item Tables~\ref{tab:bp_classification} and~\ref{tab:hr_classification} present the clinical criteria used to categorize blood pressure (BP) and heart rate (HR), respectively. Table~\ref{tab:bmi_categories} summarizes the body mass index (BMI) categories based on WHO guidelines.

\item Table~\ref{tab:liu_mapping} describes the mapping between rhythm labels used in the Liu et al. dataset and the classification tasks defined in this work. Table~\ref{tab:bp_comparison} provides a comparison of BP estimation performance with previously published studies using MIMIC-III-derived datasets.

\item Tables~\ref{tab:bp_hr_auroc} and~\ref{tab:bp_hr_threshold} report detailed rhythm classification performance across BP and HR physiological categories.

\item Tables~\ref{tab:gender_bmi_ethnicity_auroc} and~\ref{tab:aftasks_demographics_threshold} present subgroup analyses across demographic factors, including gender, BMI, and ethnicity.

\item Table~\ref{tab:ethnicity_mapping} lists the mapping of original MIMIC-III ethnicity labels into aggregated demographic groups used in the fairness and subgroup analyses.

\end{itemize}

\begin{table*}[h!]
\caption{Rhythm mapping after applying sample and patient threshold rules. \added{OTHER is not considered as a separate class, but samples are retained with zero labels for all outputs to improve robustness to non-target rhythms.}}
\renewcommand{\arraystretch}{1.1}
\label{tab:heart-rhythms_after_filter}
\centering
\begin{tabular}{l l r r l l l} 
\toprule
\textbf{Rhythm} & \textbf{Description} & \textbf{Patients} & \textbf{Samples (30s)} 
& \multicolumn{3}{c}{\textbf{Rhythm per Tasks (mapped)}} \\
\cmidrule(lr){5-7}
& & & 
& \textbf{AF (Task 1)} & \textbf{SAA (Task 2)} & \textbf{ARRH (Task 3)} \\
\midrule

SR & Sinus rhythm & 5,316 & 3,950,724 & OTHER & SINUS & SR \\
STACH & Sinus tachycardia & 2,838 & 1,192,025 & OTHER & SINUS & STACH \\
AF & Atrial fibrillation & 1,132 & 597,769 & AF & AF+AFLT & AF \\
SBRAD & Sinus bradycardia & 1,608 & 198,432 & OTHER & SINUS & SBRAD \\
VPACE & Ventricular pacing & 282 & 126,810 & OTHER & OTHER & VPACE \\
AVPACE & Atrioventricular pacing & 166 & 67,719 & OTHER & OTHER & AVPACE \\
AFLT & Atrial flutter & 210 & 44,811 & AFLT & AF+AFLT & AFLT \\
APACE & Atrial pacing & 145 & 44,415 & OTHER & OTHER & APACE \\
SARRH & Sinus arrhythmia & 105 & 7,224 & OTHER & SINUS & SARRH \\
JR & Junctional rhythm & 78 & 6,823 & OTHER & OTHER & JR \\
SVTACH & Supraventricular tachycardia & 193 & 6,342 & OTHER & OTHER & SVTACH \\
MATACH & Multifocal atrial tachycardia & 16 & 1,725 & OTHER & OTHER & MATACH \\
VTACH & Ventricular tachycardia & 53 & 1,273 & OTHER & OTHER & VTACH \\
WAPACE & Wandering atrial pacemaker & 7 & 854 & OTHER & OTHER & OTHER \\
JTACH & Junctional tachycardia & 16 & 571 & OTHER & OTHER & OTHER \\
OTHER & Other & 9 & 349 & OTHER & OTHER & OTHER \\
PATACH & Paroxysmal atrial tachycardia & 7 & 196 & OTHER & OTHER & OTHER \\
VFIB & Ventricular fibrillation & 9 & 147 & OTHER & OTHER & OTHER \\
ASYS & Asystole & 15 & 113 & OTHER & OTHER & OTHER \\
IDIOV & Idioventricular rhythm & 7 & 111 & OTHER & OTHER & OTHER \\
1AVB & 1\textsuperscript{st} degree AV block & 304 & 111,729 & exclude & exclude & exclude \\
LBBB & Left bundle branch block & 45 & 22,667 & exclude & exclude & exclude \\
RBBB & Right bundle branch block & 43 & 13,092 & exclude & exclude & exclude \\
2AVBM1 & 2\textsuperscript{nd} degree AV block Mobitz type 1 & 20 & 1,768 & exclude & exclude & exclude \\
3AVB & 3\textsuperscript{rd} degree AV block & 27 & 1,360 & exclude & exclude & exclude \\
2AVBM2 & 2\textsuperscript{nd} degree AV block Mobitz type 2 & 19 & 705 & exclude & exclude & exclude \\
\midrule
\textbf{Total} & &  & \textbf{6,399,754} &  &  &  \\
\bottomrule
\end{tabular}
\end{table*}

\begin{table*}[!htbp]
\centering
\renewcommand{\arraystretch}{1.4}  
\caption{BP classification criteria (\cite{whelton20182017, sharma2023hypotension}). Note that when systolic and diastolic blood pressure values fall into different categories, the hypertension stage is assigned according to the higher category, in accordance with established clinical guideline conventions\cite{whelton20182017}.}
\begin{tabular}{lccc}
\toprule
BP Category & \textbf{Systolic (SBP)} & \textbf{Operator} & \textbf{Diastolic (DBP)} \\
\midrule
Hypotension & $< 90$ mmHg & OR & $< 60$ mmHg \\
Normal & $90$–$119$ mmHg & AND & $60$–$79$ mmHg \\
Elevated & $120$–$129$ mmHg & AND & $< 80$ mmHg \\
Stage 1 Hypertension & $130$–$139$ mmHg & OR & $80$–$89$ mmHg \\
Stage 2 Hypertension & $\geq 140$ mmHg & OR & $\geq 90$ mmHg \\
\bottomrule
\end{tabular}
\label{tab:bp_classification}
\end{table*}

\begin{table*}[!htbp]
\centering
\renewcommand{\arraystretch}{1.4}  
\caption{HR classification criteria for adults \cite{AHA_2005_brady_tachy}. Note: sinus tachycardia is defined as a sinus node discharge rate $>100$ times per minute, i.e., HR $>100$~bpm.}
\small
\begin{tabular}{lcc}
\toprule
\textbf{HR Category} & \textbf{HR (bpm)}  \\
\midrule
Bradycardia & $<60$  \\

Normal  & $60$--$100$  \\
Tachycardia & $>100$ \\
\bottomrule
\end{tabular}
\label{tab:hr_classification}
\end{table*}

\begin{table}[t]
\centering
\caption{BMI categories according to the WHO~\cite{who2000obesity}.}
\renewcommand{\arraystretch}{1.4}
\begin{tabular}{lc}
\toprule
\textbf{Category} & \textbf{BMI (kg/m$^2$)} \\ \midrule
Underweight & $< 18.5$ \\
Normal      & $18.5$--$24.9$ \\
Overweight  & $25.0$--$29.9$ \\
Obese       & $\geq 30$ \\ \bottomrule
\end{tabular}
\label{tab:bmi_categories}
\end{table}

\begin{table*}[h]
\centering
\caption{Rhythm mapping between Liu et al. dataset \cite{liu2022multiclass} and \dbname~classification tasks}
\renewcommand{\arraystretch}{1.4}  
\label{tab:liu_mapping}
\scalebox{1.2}{ 
\begin{tabular}{lccc}
\toprule
\textbf{Liu et al. Rhythm} & \textbf{AF Task Mapping}& \textbf{SAA Task Mapping} & \textbf{ARRH Task Mapping}  \\
\midrule
0: Sinus rhythm & OTHER & SINUS & SR \\
1: Premature ventricular contraction & Excluded & Excluded & Excluded \\
2: Premature atrial contraction & Excluded & Excluded & Excluded \\
3: Ventricular tachycardia & OTHER & OTHER & VTACH \\
4: Supraventricular tachycardia & OTHER & OTHER & SVTACH \\
5: AF & AF & AF+AFLT & AF \\
\bottomrule
\end{tabular}
}
\end{table*}

\begin{table*}[htbp]
    \centering
    \caption{Comparison of BP Estimation Performance}
    \label{tab:bp_comparison}
    \scalebox{0.9}{ 
    \begin{tabular}{l c c c c l}
        \toprule
        \textbf{Study} & \textbf{Patients} (\textit{n}) & \textbf{Data Duration} (\si{\hour}) & \textbf{SBP MAE} (\si{\mmHg}) & \textbf{DBP MAE} (\si{\mmHg}) & \textbf{Data Source/Subset} \\
        \midrule
        Ours   & \textbf{2,391} & \textbf{20,572} & $\sim 16\text{--}17$ & $\sim 9\text{--}10$ &  MIMIC-III (\dbname{}) \\
        \addlinespace
        Moulaeifard et al. \cite{moulaeifard2025generalizable} & 1,474 & 2,357 & $\sim 15\text{--}17$ & $\sim 9\text{--}10$ & MIMIC-III (PulseDB \cite{wang2023pulsedb}) \\
        Sanches et al. (MIMIC-BP) \cite{sanches2024mimic} & 1,524 & 380 &   $\sim  \textbf{13\text{--}14}$ & $\sim 9\text{--}10$ & MIMIC-III (Custom Subset) \\
        \bottomrule
    \end{tabular}
    }
\end{table*}

\begin{table*}[!h]
\centering
\caption{AUROC performance of heart rhythm detection across BP and HR categories (XResNet1d101). In-distribution baselines are from Table~\ref{tab:aftasks_results}. Dashes indicate insufficient samples. HTN: hypertension; HPN: hypotension.}
\label{tab:bp_hr_auroc}
\renewcommand{\arraystretch}{1.4}
\scalebox{1.0}{
\begin{tabular}{ll|c|ccccc|ccc}
\toprule
\multirow{2}{*}{\textbf{Task}} & \multirow{2}{*}{\textbf{Rhythm}} &
\multicolumn{1}{c|}{\textbf{Baseline $\uparrow$}} &
\multicolumn{5}{c|}{\textbf{BP category $\uparrow$}} &
\multicolumn{3}{c}{\textbf{HR category $\uparrow$}} \\
& & \multicolumn{1}{c|}{} 
& \textbf{HPN} & \textbf{Normal} & \textbf{Elevated} & \textbf{Stage 1 HTN} & \multicolumn{1}{c|}{\textbf{Stage 2 HTN}}
& \textbf{Bradycardia} & \textbf{Normal} & \textbf{Tachycardia} \\
\midrule

\multirow{2}{*}{\rotatebox{90}{AF}}
& AF    & 0.96 & 0.95 & 0.99 & 0.99 & 0.99 & 0.97 & 0.88 & 0.96 & 0.97 \\
& AFLT  & 0.77 & 0.80 & 0.77 & 0.78 & 0.80 & 0.84 & 0.74 & 0.76 & 0.86 \\

\midrule

\multirow{2}{*}{\rotatebox{90}{SAA}}
& SINUS        & 0.92 & 0.88 & 0.95 & 0.96 & 0.97 & 0.96 & 0.84 & 0.91 & 0.96 \\
& AF+AFLT    & 0.95 & 0.96 & 0.99 & 0.99 & 0.99 & 0.97 & 0.87 & 0.95 & 0.97 \\

\midrule

\multirow{13}{*}{\rotatebox{90}{ARRH}}
& SR      & 0.93 & 0.92 & 0.95 & 0.96 & 0.96 & 0.94 & 0.72 & 0.86 & 0.76 \\
& STACH   & 0.97 & 0.97 & 0.96 & 0.97 & 0.98 & 0.98 & 0.72 & 0.90 & 0.88 \\
& AF      & 0.96 & 0.95 & 0.99 & 0.99 & 0.99 & 0.97 & 0.87 & 0.96 & 0.97 \\
& SBRAD   & 0.96 & 0.97 & 0.96 & 0.97 & 0.98 & 0.97 & 0.75 & 0.90 & 0.58 \\
& VPACE   & 0.79 & 0.73 & 0.80 & 0.91 & 0.82 & 0.65 & 0.82 & 0.77 & 0.72 \\
& AVPACE  & 0.89 & 0.82 & 0.96 & 0.99 & 0.96 & 0.97 & 0.91 & 0.89 & 0.82 \\
& AFLT    & 0.78 & 0.87 & 0.81 & 0.89 & 0.89 & 0.84 & 0.83 & 0.76 & 0.83 \\
& APACE   & 0.80 & 0.73 & 0.79 & 0.78 & 0.87 & 0.76 & 0.61 & 0.80 & 0.96 \\
& SARRH   & 0.61 & 0.89 & 0.78 & 0.86 & 0.90 & 0.85 & 0.93 & 0.80 & 0.73 \\
& JR      & 0.62 & 0.47 & 0.42 & 0.81 & --   & 0.62 & 0.86 & 0.53 & 0.63 \\
& SVTACH  & 0.78 & 0.93 & 0.62 & --   & 0.47 & 0.86 & 0.88 & 0.80 & 0.73 \\
& MATACH  & 0.84 & 0.94 & --   & 0.84 & --   & 0.87 & --   & 0.81 & 0.88 \\
& VTACH   & 0.75 & 0.92 & 0.92 & --   & --   & 0.42 & 0.69 & 0.83 & 0.62 \\

\bottomrule
\end{tabular}}
\end{table*}

\begin{table*}[t]
\centering
\caption{\added{Sensitivity (at specificity $\geq$ 0.8)\,and \,specificity (at sensitivity $\geq$ 0.8) across BP and HR categories (XResNet1d101). Values are reported as sensitivity\,/\,specificity. Baseline results are from Table~\ref{tab:aftasks_results}. Dashes indicate insufficient samples.}}
\label{tab:bp_hr_threshold}
\renewcommand{\arraystretch}{1.4}
\scalebox{0.9}{
\begin{tabular}{ll|c|ccccc|ccc}
\toprule
\multirow{2}{*}{\textbf{Task}} & \multirow{2}{*}{\textbf{Rhythm}} &
\multicolumn{1}{c|}{\textbf{Baseline $\uparrow$}} &
\multicolumn{5}{c|}{\textbf{BP category $\uparrow$}} &
\multicolumn{3}{c}{\textbf{HR category $\uparrow$}} \\
& & \multicolumn{1}{c|}{} 
& \textbf{HPN} & \textbf{Normal} & \textbf{Elevated} & \textbf{Stage 1 HTN} & \multicolumn{1}{c|}{\textbf{Stage 2 HTN}}
& \textbf{Bradycardia} & \textbf{Normal} & \textbf{Tachycardia} \\
\midrule

\multirow{2}{*}{\rotatebox{90}{AF}}
& AF    & 0.92 / 0.98 & 0.92 / 0.98 & 0.99 / 0.99 & 0.99 / 0.99 & 0.99 / 0.99 & 0.94 / 0.99 & 0.79 / 0.78 & 0.93 / 0.99 & 0.96 / 0.97 \\
& AFLT  & 0.62 / 0.61 & 0.70 / 0.73 & 0.61 / 0.65 & 0.54 / 0.70 & 0.63 / 0.72 & 0.78 / 0.79 & 0.51 / 0.50 & 0.62 / 0.59 & 0.73 / 0.76 \\

\midrule

\multirow{2}{*}{\rotatebox{90}{SAA}}
& SINUS      & 0.89 / 0.86 & 0.77 / 0.78 & 0.97 / 0.91 & 1.00 / 0.95 & 0.99 / 0.96 & 0.99 / 0.93 & 0.70 / 0.74 & 0.87 / 0.84 & 0.97 / 0.95 \\
& AF+AFLT  & 0.93 / 0.98 & 0.92 / 0.98 & 0.99 / 0.99 & 0.99 / 1.00 & 0.99 / 1.00 & 0.95 / 0.99 & 0.80 / 0.80 & 0.92 / 0.98 & 0.96 / 0.98 \\

\midrule

\multirow{13}{*}{\rotatebox{90}{ARRH}}
& SR     & 0.93 / 0.89 & 0.93 / 0.88 & 0.95 / 0.94 & 0.97 / 0.95 & 0.96 / 0.95 & 0.94 / 0.92 & 0.51 / 0.49 & 0.75 / 0.76 & 0.58 / 0.58 \\
& STACH  & 0.97 / 0.97 & 0.97 / 0.98 & 0.96 / 0.97 & 0.97 / 0.98 & 0.97 / 0.98 & 0.98 / 0.97 & 0.67 / 0.13 & 0.85 / 0.85 & 0.80 / 0.78 \\
& AF     & 0.94 / 0.96 & 0.92 / 0.98 & 0.99 / 0.99 & 0.99 / 1.00 & 0.99 / 1.00 & 0.96 / 0.98 & 0.80 / 0.80 & 0.93 / 0.98 & 0.96 / 0.97 \\
& SBRAD  & 0.95 / 0.98 & 0.94 / 0.99 & 0.94 / 0.99 & 0.95 / 0.99 & 0.98 / 0.99 & 0.96 / 0.99 & 0.52 / 0.55 & 0.87 / 0.88 & 0.27 / 0.29 \\
& VPACE  & 0.62 / 0.60 & 0.47 / 0.51 & 0.61 / 0.65 & 0.86 / 0.85 & 0.61 / 0.65 & 0.29 / 0.44 & 0.69 / 0.70 & 0.59 / 0.56 & 0.47 / 0.53 \\
& AVPACE & 0.81 / 0.81 & 0.73 / 0.69 & 0.94 / 0.95 & 1.00 / 0.99 & 0.98 / 0.95 & 0.96 / 0.96 & 0.89 / 0.86 & 0.79 / 0.79 & 0.61 / 0.60 \\
& AFLT   & 0.65 / 0.60 & 0.79 / 0.79 & 0.57 / 0.71 & 0.85 / 0.82 & 0.85 / 0.84 & 0.77 / 0.79 & 0.55 / 0.67 & 0.60 / 0.63 & 0.64 / 0.60 \\
& APACE  & 0.63 / 0.61 & 0.45 / 0.54 & 0.68 / 0.53 & 0.66 / 0.53 & 0.78 / 0.76 & 0.45 / 0.60 & 0.22 / 0.44 & 0.66 / 0.63 & 0.97 / 0.96 \\
& SARRH  & 0.40 / 0.26 & 0.78 / 0.79 & 0.57 / 0.61 & 0.70 / 0.78 & 0.80 / 0.82 & 0.68 / 0.75 & 0.92 / 0.90 & 0.57 / 0.60 & 0.38 / 0.59 \\
& JR     & 0.45 / 0.20 & 0.14 / 0.12 & 0.19 / 0.13 & 0.64 / 0.78 & --          & 0.25 / 0.47 & 0.74 / 0.77 & 0.35 / 0.11 & 0.33 / 0.33 \\
& SVTACH & 0.65 / 0.54 & 0.96 / 0.87 & 0.42 / 0.34 & --          & 0.03 / 0.33 & 0.69 / 0.75 & 0.78 / 0.80 & 0.57 / 0.63 & 0.97 / 0.96 \\
& MATACH & 0.72 / 0.72 & 0.83 / 0.90 & --          & 1.00 / 0.84 & --          & 1.00 / 0.87 & --          & 0.65 / 0.68 & 0.80 / 0.78 \\
& VTACH  & 0.60 / 0.49 & 0.90 / 0.86 & 0.97 / 0.89 & --          & --          & 0.15 / 0.19 & 0.62 / 0.40 & 0.67 / 0.69 & 0.52 / 0.25 \\

\bottomrule
\end{tabular}}
\end{table*}

\begin{table*}[h]
\centering
\caption{Performance (AUROC) across gender, BMI, and ethnicity categories (XResNet1D101). Baseline results are taken from Table~\ref{tab:aftasks_results}. Dashes indicate insufficient samples. The "Underweight" BMI category was excluded due to the insufficient number of samples.}
\label{tab:gender_bmi_ethnicity_auroc}
\renewcommand{\arraystretch}{1.3}
\scalebox{1.0}{
\begin{tabular}{ll|c|cc|ccc|cccc}
\toprule
\multirow{2}{*}{\textbf{Task}} &
\multirow{2}{*}{\textbf{Rhythm}} &
\multicolumn{1}{c|}{\textbf{Baseline $\uparrow$}} &
\multicolumn{2}{c|}{\textbf{Gender $\uparrow$}} &
\multicolumn{3}{c|}{\textbf{BMI $\uparrow$}} &
\multicolumn{4}{c}{\textbf{Ethnicity $\uparrow$}} \\
& &
& Female & Male &
Normal & Overweight & Obese &
White & Black & Asian & Hispanic \\
\midrule

\multirow{2}{*}{\rotatebox{90}{AF}}
& AF    & 0.96 & 0.95 & 0.96 & 0.92 & 0.92 & 0.98 & 0.95 & 0.96 & 0.98 & 0.95 \\
& AFLT  & 0.77 & 0.84 & 0.75 & 0.72 & 0.75 & 0.75 & 0.74 & 0.93 & -- & -- \\
\midrule

\multirow{2}{*}{\rotatebox{90}{SAA}}
& SINUS        & 0.92 & 0.91 & 0.93 & 0.84 & 0.93 & 0.96 & 0.93 & 0.93 & 0.98 & 0.82 \\
& AF + AFLT  & 0.95 & 0.95 & 0.95 & 0.93 & 0.93 & 0.97 & 0.95 & 0.95 & 0.98 & 0.95 \\

\midrule

\multirow{13}{*}{\rotatebox{90}{ARRH}}
& SR      & 0.93 & 0.93 & 0.92 & 0.87 & 0.92 & 0.95 & 0.93 & 0.95 & 0.98 & 0.91 \\
& STACH   & 0.97 & 0.97 & 0.97 & 0.96 & 0.98 & 0.97 & 0.97 & 0.97 & 0.99 & 0.98 \\
& AF      & 0.96 & 0.95 & 0.96 & 0.94 & 0.93 & 0.97 & 0.95 & 0.97 & 0.98 & 0.95 \\
& SBRAD   & 0.96 & 0.96 & 0.96 & 0.98 & 0.96 & 0.95 & 0.97 & 0.97 & 0.89 & 0.91 \\
& VPACE   & 0.79 & 0.71 & 0.84 & 0.65 & 0.88 & 0.95 & 0.86 & --   & 0.97 & 0.95 \\
& AVPACE  & 0.89 & 0.85 & 0.91 & 0.82 & 0.91 & 0.89 & 0.90 & --   & --   & 0.85 \\
& AFLT    & 0.78 & 0.84 & 0.77 & 0.76 & 0.77 & 0.79 & 0.76 & 0.88 & --   & -- \\
& APACE   & 0.80 & 0.87 & 0.73 & 0.86 & 0.71 & 0.82 & 0.80 & --   & --   & 0.63 \\
& SARRH   & 0.61 & 0.56 & 0.78 & 0.80 & 0.91 & 0.53 & 0.50 & 0.88 & --   & 0.84 \\
& JR      & 0.62 & 0.75 & 0.59 & 0.18 & 0.42 & 0.82 & 0.45 & 0.96 & --   & -- \\
& SVTACH  & 0.78 & 0.80 & 0.73 & 0.80 & 0.64 & 0.93 & 0.73 & 0.82 & 0.97 & 0.94 \\
& MATACH  & 0.84 & 0.93 & 0.75 & 0.90 & 0.94 & 0.84 & 0.83 & --   & --   & -- \\
& VTACH   & 0.75 & 0.86 & 0.66 & 0.80 & 0.80 & --   & 0.84 & 0.39 & 0.74 & 0.79 \\

\bottomrule
\end{tabular}}
\end{table*}

\begin{table*}[h]
\centering
\caption{\added{Sensitivity  (at specificity $\geq$ 0.8\,) and \,specificity (at sensitivity $\geq$ 0.8) across demographic groups (XResNet1d101). Values are reported as sensitivity\,/\,specificity. Baseline results are from Table~\ref{tab:aftasks_results}. Dashes indicate insufficient samples.}}
\label{tab:aftasks_demographics_threshold}
\renewcommand{\arraystretch}{1.3}
\scalebox{0.9}{
\begin{tabular}{ll|c|cc|ccc|cccc}
\toprule
\multirow{2}{*}{\textbf{Task}} &
\multirow{2}{*}{\textbf{Rhythm}} &
\multicolumn{1}{c|}{\textbf{Baseline $\uparrow$}} &
\multicolumn{2}{c|}{\textbf{Gender $\uparrow$}} &
\multicolumn{3}{c|}{\textbf{BMI $\uparrow$}} &
\multicolumn{4}{c}{\textbf{Ethnicity $\uparrow$}} \\
& &
& Female & Male &
Normal & Overweight & Obese &
White & Black & Asian & Hispanic \\
\midrule

\multirow{2}{*}{\rotatebox{90}{AF}}
& AF    & 0.92 / 0.98 & 0.92 / 0.98 & 0.93 / 0.98 & 0.89 / 0.94 & 0.87 / 0.92 & 0.97 / 0.99 & 0.90 / 0.97 & 0.95 / 0.98 & 0.96 / 0.99 & 0.93 \\
& AFLT  & 0.62 / 0.61 & 0.74 / 0.74 & 0.60 / 0.59 & 0.49 / 0.53 & 0.54 / 0.60 & 0.70 / 0.52 & 0.92 / 0.58 & 0.89 / 0.91 & -- & -- \\

\midrule

\multirow{2}{*}{\rotatebox{90}{SAA}}
& SINUS      
& 0.89\,/\,0.86
& 0.86\,/\,0.84 & 0.91\,/\,0.88
& 0.69\,/\,0.72 & 0.92\,/\,0.88 & 0.97\,/\,0.94
& 0.93\,/\,0.89 & 0.93\,/\,0.91 & 0.99\,/\,0.98 & 0.66\,/\,0.62 \\

& AF+AFLT  
& 0.93\,/\,0.98
& 0.93\,/\,0.98 & 0.93\,/\,0.98
& 0.90\,/\,0.94 & 0.88\,/\,0.93 & 0.96\,/\,0.99
& 0.92\,/\,0.97 & 0.93\,/\,0.96 & 0.97\,/\,0.99 & 0.91\,/\,0.91 \\

\midrule

\multirow{13}{*}{\rotatebox{90}{ARRH}}

& SR
& 0.93\,/\,0.89
& 0.93\,/\,0.90 & 0.92\,/\,0.89
& 0.80\,/\,0.80 & 0.92\,/\,0.88 & 0.96\,/\,0.94
& 0.93\,/\,0.90 & 0.96\,/\,0.93 & 0.99\,/\,0.98 & 0.88\,/\,0.84 \\

& STACH
& 0.97\,/\,0.97
& 0.96\,/\,0.97 & 0.97\,/\,0.98
& 0.95\,/\,0.95 & 0.97\,/\,0.98 & 0.97\,/\,0.89
& 0.96\,/\,0.97 & 0.97\,/\,0.98 & 0.99\,/\,0.99 & 0.98\,/\,0.98 \\

& AF
& 0.94\,/\,0.96
& 0.93\,/\,0.97 & 0.94\,/\,0.97
& 0.91\,/\,0.94 & 0.88\,/\,0.92 & 0.97\,/\,0.98
& 0.93\,/\,0.97 & 0.97\,/\,0.97 & 0.96\,/\,0.99 & 0.93\,/\,0.93 \\

& SBRAD
& 0.95\,/\,0.98
& 0.94\,/\,0.98 & 0.95\,/\,0.98
& 0.96\,/\,0.99 & 0.94\,/\,0.98 & 0.92\,/\,0.99
& 0.95\,/\,0.98 & 0.97\,/\,0.98 & 0.88\,/\,0.92 & 0.85\,/\,0.95 \\

& VPACE
& 0.62\,/\,0.60
& 0.46\,/\,0.48 & 0.73\,/\,0.73
& 0.37\,/\,0.39 & 0.83\,/\,0.84 & 0.94\,/\,0.92
& 0.77\,/\,0.78 & -- & 0.98\,/\,0.93 & 0.98\,/\,0.92 \\

& AVPACE
& 0.81\,/\,0.81
& 0.73\,/\,0.73 & 0.86\,/\,0.88
& 0.67\,/\,0.69 & 0.87\,/\,0.89 & 0.85\,/\,0.88
& 0.85\,/\,0.86 & -- & -- & 0.69\,/\,0.71 \\

& AFLT
& 0.65\,/\,0.60
& 0.76\,/\,0.75 & 0.63\,/\,0.62
& 0.60\,/\,0.63 & 0.59\,/\,0.62 & 0.70\,/\,0.69
& 0.61\,/\,0.61 & 0.84\,/\,0.84 & -- & -- \\

& APACE
& 0.63\,/\,0.61
& 0.78\,/\,0.78 & 0.49\,/\,0.52
& 0.78\,/\,0.78 & 0.36\,/\,0.56 & 0.65\,/\,0.54
& 0.65\,/\,0.63 & -- & -- & 0.19\,/\,0.49 \\

& SARRH
& 0.40\,/\,0.26
& 0.33\,/\,0.22 & 0.64\,/\,0.51
& 0.62\,/\,0.64 & 0.89\,/\,0.86 & 0.23\,/\,0.28
& 0.26\,/\,0.21 & 0.77\,/\,0.77 & -- & 0.64\,/\,0.75 \\

& JR
& 0.45\,/\,0.20
& 0.65\,/\,0.43 & 0.40\,/\,0.19
& 0.00\,/\,0.06 & 0.16\,/\,0.13 & 0.73\,/\,0.65
& 0.22\,/\,0.13 & 0.95\,/\,0.94 & -- & -- \\

& SVTACH
& 0.65\,/\,0.54
& 0.71\,/\,0.67 & 0.53\,/\,0.43
& 0.65\,/\,0.67 & 0.45\,/\,0.29 & 0.97\,/\,0.89
& 0.57\,/\,0.46 & 0.69\,/\,0.64 & 1.00\,/\,0.96 & 0.93\,/\,0.95 \\

& MATACH
& 0.72\,/\,0.72
& 0.94\,/\,0.90 & 0.49\,/\,0.60
& 0.93\,/\,0.87 & 0.91\,/\,0.87 & 0.72\,/\,0.75
& 0.70\,/\,0.71 & -- & -- & -- \\

& VTACH
& 0.60\,/\,0.49
& 0.70\,/\,0.73 & 0.53\,/\,0.29
& 0.49\,/\,0.63 & 0.72\,/\,0.71 & --
& 0.77\,/\,0.76 & 0.22\,/\,0.16 & 0.50\,/\,0.51 & 0.60\,/\,0.61 \\

\bottomrule
\end{tabular}}
\end{table*}

\begin{table*}[ht]
\centering
\caption{\added{Mapping of original MIMIC-III ethnicity labels to aggregated ethnicity groups, based on
        U.S. Office of Management and Budget Statistical Policy Directive 
        No.~15 (1997) \cite{office1997revisions}. It is worth noting that due to extremely small numbers of subjects in certain categories (e.g., Caribbean Island, $n = 1$; American Indian/Alaska Native, $n = 2$; Native Hawaiian/Pacific Islander, $n = 2$; multiracial, $n = 17$), these groups were consolidated into the “OTHER” category to ensure statistical stability of subgroup analyses.}}
\label{tab:ethnicity_mapping}
\renewcommand{\arraystretch}{1.2}
\begin{tabular}{ll}
\toprule
\textbf{Original Ethnicity Label} & \textbf{Aggregated Group} \\
\midrule
WHITE & White \\
WHITE - RUSSIAN & White \\
WHITE - EASTERN EUROPEAN & White \\
WHITE - OTHER EUROPEAN & White \\
WHITE - BRAZILIAN & White \\
PORTUGUESE & White \\
MIDDLE EASTERN & White \\
\midrule
BLACK/AFRICAN AMERICAN & Black \\
BLACK/AFRICAN & Black \\
BLACK/CAPE VERDEAN & Black \\
BLACK/HAITIAN & Black \\

\midrule
ASIAN & Asian \\
ASIAN - CHINESE & Asian \\
ASIAN - ASIAN INDIAN & Asian \\
ASIAN - CAMBODIAN & Asian \\
ASIAN - VIETNAMESE & Asian \\
ASIAN - OTHER & Asian \\
ASIAN - FILIPINO & Asian \\
ASIAN - KOREAN & Asian \\
ASIAN - JAPANESE & Asian \\
\midrule

HISPANIC OR LATINO & Hispanic \\
HISPANIC/LATINO - PUERTO RICAN & Hispanic \\
HISPANIC/LATINO - DOMINICAN & Hispanic \\
HISPANIC/LATINO - GUATEMALAN & Hispanic \\
HISPANIC/LATINO - CENTRAL AMERICAN (OTHER) & Hispanic \\
HISPANIC/LATINO - SALVADORAN & Hispanic \\
HISPANIC/LATINO - MEXICAN & Hispanic \\
HISPANIC/LATINO - COLOMBIAN & Hispanic \\
HISPANIC/LATINO - HONDURAN & Hispanic \\
HISPANIC/LATINO - CUBAN & Hispanic \\

\midrule
OTHER & Other \\
UNKNOWN/NOT SPECIFIED & Other \\
UNABLE TO OBTAIN & Other \\
PATIENT DECLINED TO ANSWER & Other \\
AMERICAN INDIAN/ALASKA NATIVE & Other \\
NATIVE HAWAIIAN OR OTHER PACIFIC ISLANDER & Other \\
MULTI RACE ETHNICITY & Other \\
CARIBBEAN ISLAND & Other \\
\bottomrule
\end{tabular}
\end{table*}

\end{document}